\documentclass{article}
\usepackage{spconf,amsmath,graphicx}
\usepackage{algorithmic}
\usepackage[ruled,vlined]{algorithm2e}
\usepackage{textcomp}
\usepackage{amssymb}
\usepackage{xcolor}
\usepackage{mathtools}
\usepackage{cite}
\usepackage{multirow}
\usepackage{mathrsfs}
\usepackage{scalerel}
\usepackage{hyperref}
\input{my_symbol.sty}

\title{Multi-Source Domain Adaptation with Transformer-based Feature Generation for Subject-Independent EEG-based Emotion Recognition}
\name{Shadi Sartipi$^{\star}$, and Mujdat Cetin$^{\star}$$^{\dagger}$\thanks{This work has been partially supported by the National Science Foundation (NSF) under grants CCF-1934962 and DGE-1922591, and by the BRAIN Initiative of the National Institutes of Health through grant U19NS128613.}}
\address{$^{\star}$Department of Electrical and Computer Engineering, University of Rochester, Rochester, NY, USA\\
$^{\dagger}$Goergen Institute for Data Science, University of Rochester, Rochester, NY, USA}
\begin{document}
%
\maketitle
\begin{abstract}
Although deep learning-based algorithms have demonstrated excellent performance in automated emotion recognition via electroencephalogram (EEG) signals, variations across brain signal patterns of individuals can diminish the model's effectiveness when applied across different subjects. While transfer learning techniques have exhibited promising outcomes, they still encounter challenges related to inadequate feature representations and may overlook the fact that source subjects themselves can possess distinct characteristics. In this work, we propose a multi-source domain adaptation approach with a transformer-based feature generator (MSDA-TF) designed to leverage information from multiple sources. The proposed feature generator retains convolutional layers to capture shallow spatial, temporal, and spectral EEG data representations, while self-attention mechanisms extract global dependencies within these features. During the adaptation process, we group the source subjects based on correlation values and aim to align the moments of the target subject with each source as well as within the sources. MSDA-TF is validated on the SEED dataset and is shown to yield promising results.
\end{abstract}
\section{Introduction}
\label{sec:intro}
Affective computing intends to process, handle, identify, and react to individuals' emotional states. It holds great potential across various application areas ranging from healthcare and education to brain-computer interfaces (BCIs) \cite{shih2012brain}. EEG-based emotion recognition has gained great attention due to the high temporal resolution, data adequacy, and clear response to emotional stimuli \cite{zheng2017identifying}. While various studies try to capture the time and frequency features from the EEG data for emotion recognition, the high subject dependency of the EEG data prevents getting the desired performance \cite{samek2013transferring}. This variability across different subjects could be due to head shape, mental states, noise, etc \cite{samek2013transferring}.

Deep learning approaches have been applied widely in this domain to find the features that can discriminate the emotional states \cite{schirrmeister2017deep}. EEGNet \cite{lawhern2018eegnet} and ConvNet \cite{schirrmeister2017deep} are two convolutional neural networks (CNN) based architectures that showed great performance. Alongside the spatial information, the temporal dependencies can also boost the model's performance. One approach is using CNN and long-short-term memory (LSTM) networks to capture the spatial and temporal features \cite{sartipi2023hybrid}. Transformers (TF) are also utilized to capture the long-term dependencies \cite{liu2021swin}. However, there is still room to find a network that can extract discriminative features across different subjects. 

The traditional approach to addressing the mentioned limitation involves using a sufficient amount of labeled target domain data (i.e., training data from the subject of interest) to calibrate the learned model, which can be time-consuming. Transfer learning, as discussed in \cite{pan2009survey}, is a common strategy applied to tackle this issue. Domain adaptation (DA) is a branch of transfer learning that employs specific metrics to enhance the performance of the target domain by minimizing domain shifts between the target and source domains. Maximum mean discrepancy
(MMD) \cite{gretton2006kernel} is a widely used metric that reduces the distance between two distributions \cite{amirshahi2022m2d2}. 
Adversarial discriminative domain adaptation (ADDA) \cite{tzeng2017adversarial}  learns a discriminative representation from source domain labels and then maps the target data to the same space through an asymmetric mapping using a domain-adversarial loss. Yet, differences among subjects in the source domain can challenge the model's learning process.

In this paper, we propose a multi-source domain adaptation approach with a transformer-based feature generator (MSDA-TF). The proposed feature generator uses CNN blocks to initially capture the local spatial, temporal, and spectral characteristics of the EEG data. Since CNN can only capture local information, the TF is applied to extract global features to compensate for the limitation of CNN. To enhance the model's performance across different subjects, we aim to align the moments of the feature distribution of multiple subjects (labeled source domains) with the test subject (unlabeled target domain). The main contributions of this paper can be summarized as follows:
\begin{itemize}
    \item A novel feature generator is proposed to capture local features via CNN, followed by the application of TF to extract global dependencies.
    \item Training subjects are grouped based on a correlation metric to form multiple source domains, and moment matching MSDA is applied to improve target domain performance.
\end{itemize}

\begin{figure}[t!]
    \centering
    \vstretch{.9}{\includegraphics[scale=0.5, angle=-90]{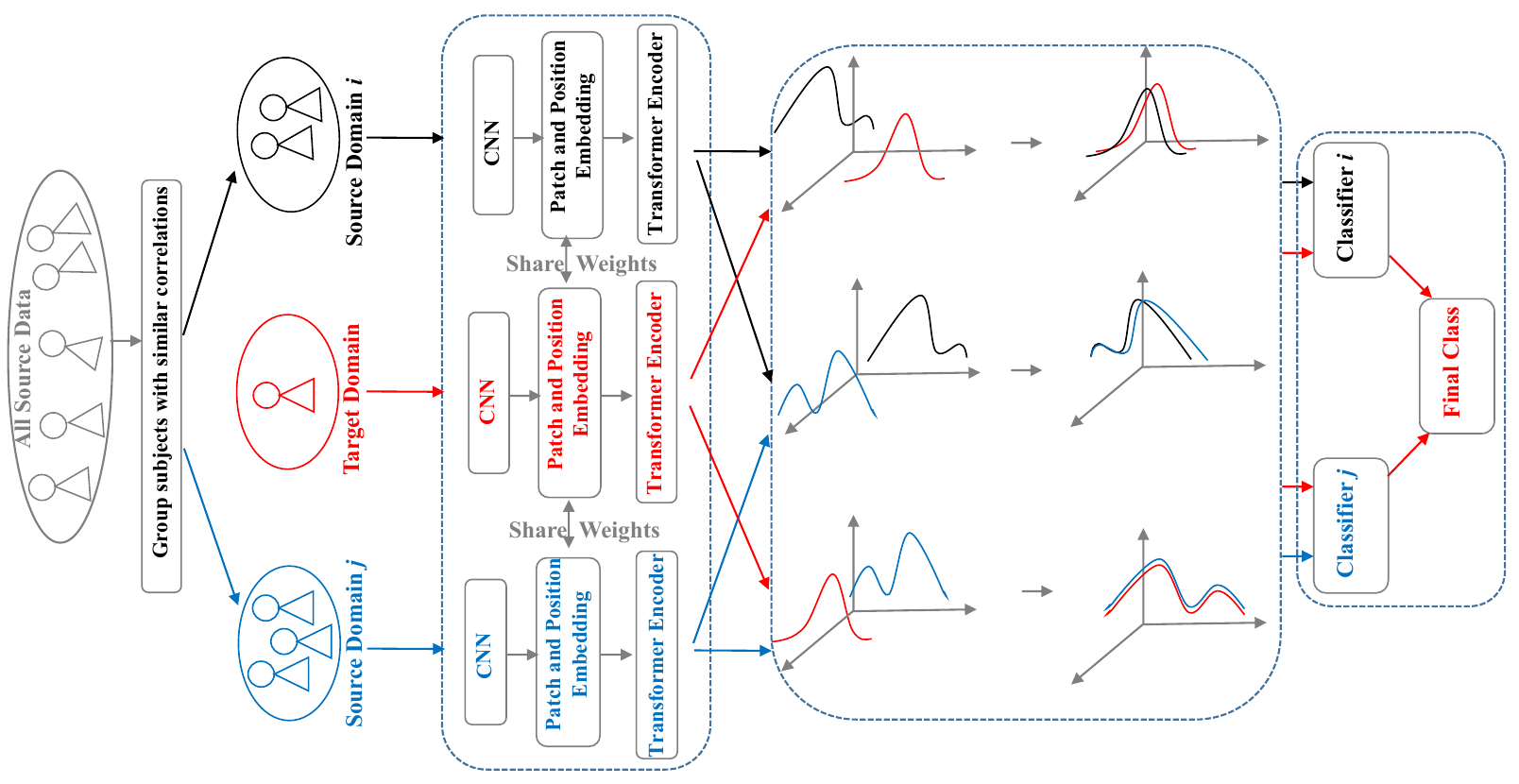}}
    \caption{Overview of the proposed MSDA-TF.}
    \label{fig:overall}
\end{figure}

\begin{table}[t]
\centering
\caption{Details of the CNN module. Conv is $2$D CNN. Parameters: Conv (number of filters; filter size) with ReLU activation and batch normalization. Maxpool (kernel size; stride). Dropout (dropout rate).}
\begin{tabular}{lc}
\textbf{Block}& \textbf{Details} \\
\hline
\multirow{3}{*}{C1}& Conv ($64$; $3$),Conv ($64$; $3$),Conv ($128$; $3$)\\
&Maxpool ($2$; $2$), Dropout ($0.30$)\\
\hline
\multirow{3}{*}{C2}& Conv ($128$; $3$),Conv ($256$; $3$),Conv ($512$; $3$)\\
&Maxpool ($2$; $2$), Dropout ($0.20$)\\
\hline\\
\end{tabular}
\label{table:cnn}
\end{table}

\section{Method}
\label{sec:method}
In this section, we describe the proposed feature generator and the domain adaptation process. The feature generator aims to learn the discriminative features, while domain adaptation d Figure~\ref{fig:overall} illustrates the overall structure of the proposed method.
\subsection{Feature Generator}
The main step in EEG emotion recognition studies is finding the features that can discriminate between different emotional states. Our feature generator consists of two main parts: the CNN module and the transformer encoder. The CNN module adopts multiple 2D convolutions. The CNN module has been utilized for its capability to extract effective features \cite{dose2018end} and reduce the dimensionality of the EEG data. This module considers the spatial, temporal, and spectral features of the EEG data. As shown in Table~\ref{table:cnn}, the CNN module comprises two blocks, C1 and C2, each containing three CNN layers, one max-pooling layer, and one dropout layer.

Following the CNN module, the extracted features are directed into the TF encoder. The architecture of TF is drawn from \cite{vaswani2017attention}, renowned for its efficacy in natural language processing. Let $H$, $W$, and $C$ represent the height, width, and channels of the CNN-module output, respectively. Initially, the TF divides the data into $n$ patches with a lower dimensionality of size $p$, where $n=\frac{H}{p}\times{}\frac{W}{p}$. Subsequently, a linear layer is applied to each patch, projecting it into a $D$-dimensional space. To uphold the inherent spatial arrangement among these patches, positional embeddings are introduced to the patch embeddings. In this study, we set $p$ and $D$ to $3$ and $64$, respectively.

The TF encoder consists of $l$ consecutive blocks of multi-head self-attention (MHA) and multi-layer perception (MLP). Each MHA block incorporates $h$ self-attention heads, with each head producing an $n\times{}d$ sequence. To perform the attention mechanism, the input vector is multiplied by three distinct weight matrices, resulting in the derivation of the query vector ($\mathrm{Q}$), key vector ($\mathrm{K}$), and value vector ($\mathrm{V}$), where $\mathrm{Q},~\mathrm{K}$, and $\mathrm{V}\in ^{n\times{}d}$. For the attention mechanism, each query vector is compared to a set of key vectors. The outcome is normalized using a softmax function and then multiplied by a set of value vectors as follows \cite{vaswani2017attention}:
\vspace{-1 pt}
\begin{equation}
\text{Attention}(\mathrm{Q},\mathrm{K},\mathrm{V})=\text{Softmax}\left(\frac{\mathrm{Q}\mathrm{K}^{\text{T}}}{\sqrt{d}}\right)\mathrm{V}
\end{equation}
\vspace{-1 pt}
To obtain the MHA, the resulting sequences from each block are concatenated into an $n \times{}dh$ sequence. In this study, we set $l$ and $h$ to $8$. Besides the MHA, the encoder also contains two MLP blocks with the number of units set to $2048$ and $1024$, respectively. The outputs of the encoder are then fed into the Softmax classifier for classification. 
\vspace{-2 pt}
\begin{algorithm}[t]\label{A:algorithm1}
  \SetAlgoLined
  \textbf{Input:} Source domains $D_s$, number of epochs $t$, and number of source domains $K$.\\
  Target domain $D_T$.\\
 \textbf{Prepare the source domains}\\
 Apply Pearson Correlation and group similar subjects.\\ 
\textbf{Training Phase}\\
    \For{$i=1,\cdots,t$}{
    \textbf{step 1:} Optimize $\mathcal{F}$ and $\mathscr{C}$ via Equation~\ref{eq:step1}\\
    \textbf{step 2:} 
    \vspace{-1 pt}
    $\min_{\mathscr{C}}\displaystyle\sum_{i=1}^{K}\mathcal{L}_{(Xsi,Ysi)}-\displaystyle\sum_{i=1}^{K}|\mathcal{C}_{i}({D}_{T})-\mathcal{C}_{i}^{\prime}({D}_{T})|$\\
    \textbf{step 3:}
    $\min_{\mathcal{F}}\displaystyle\sum_{i=1}^{K}|\mathcal{C}_{i}({D}_{T})-\mathcal{C}_{i}^{\prime}({D}_{T})|$\\
    \vspace{-5 pt}
    }

  \caption{Domain Adaptation}
\end{algorithm}
\subsection{Domain Adaptation}
Variations across subjects in a broad training set can reduce the effectiveness of transfer learning based on that set. Splitting that data into more homogeneous multiple source domains enables better domain adaptation between these source domains and the target domain. Let $D_s =\{(Z_{s1},Y_{s1}),(Z_{s2},Y_{s2}),\dots,(Z_{sK},Y_{sK})\}$ be the $k$ sets of source domain data and their labels, and $D_T=\{Z_{t}\}_{i=1}^{n}$ represent the target data without labels. Since we are dealing with more than one source domain, during the adaptation process the model aims to align the source domains with the target domain while concurrently aligning the source domains with each other by using the paradigm presented in \cite{saito2018maximum, peng2019moment}. Considering $\mathbb{E}({X}^p)=m^{(-1)}\displaystyle\sum_{i=1}^{m}x_i^p$ the $p$-order moment of $X$ with the total number of $m$ samples, to calculate the distribution differences among domains, the moment distance (MD) based on \cite{peng2019moment} is defined as follows.
\vspace{-1 pt}
\begin{multline}
\label{eq:md}
    \text{MD}(D_s,D_T)=\displaystyle\sum_{p=1}^{P} (\frac{1}{K}\displaystyle\sum_{i=1}^{K} \| \mathbb{E}({X_{si}^p})-\mathbb{E}({X_{T}^p}) \|_{2}\\
    +\frac{2!(K-2)!}{K!}\displaystyle\sum_{i=1}^{K-1}\displaystyle\sum_{j=i+1}^{K}\| \mathbb{E}({X_{si}^p})-\mathbb{E}({X_{sj}^p}) \|_{2})
\end{multline}

As shown in Figure~\ref{fig:overall}, the model consists of the feature generator, $\mathcal{F}$, and $K$ classifiers, $\mathcal{C}_{K}$ for $K$ source domains. Thus, the objective function would be a combination of the Softmax cross-entropy loss, $\mathcal{L}_{(X_{si},Y_{si})}$ for training the $K$ classifiers and the feature generator cost function, i.e., ~\eqref{eq:md}, as follows \cite{peng2019moment}.
\vspace{-2 pt}
\begin{equation}
\label{eq:step1}
\min_{\mathcal{F},\mathcal{C}}\displaystyle\sum_{i=1}^{K}\mathcal{L}_{(Xsi,Ysi)}+\lambda \min_{\mathcal{F}} \text{MD}(D_s,D_T)
\end{equation}
\vspace{-2 pt}
where $\lambda$ is the hyperparameter setting the relative weighting of the two different loss functions.
\subsection{Learning Process}
As mentioned previously, the brain responses to the same emotional state vary across different subjects. Thus, instead of considering all training subjects as a single source domain, we consider $K$ different source domains. To quantify the degree of similarity among brain responses, we apply the Pearson correlation across all training subjects without considering the labels of the data. Then, we group them into $K$ groups based on the correlation values. In this work, we set $K$ to $4$. 

During the domain adaptation process, inspired by \cite{saito2018maximum} we follow the steps presented in Algorithm~\ref{A:algorithm1}. For each source domain, let $\mathscr{C}=\{(\mathcal{C}_{i},\mathcal{C}_{i}^{\prime})\}_{i=1}^{K}$ be the pair of classifiers.  The goal of the paired classifiers is to get the target samples away from the support of the source. First, we train the feature generator and the classifier, $\mathcal{F}$ and $\mathscr{C}$, to classify the source domain samples. Second, with fixed $\mathcal{F}$, $\mathcal{C}_{i}$ and $\mathcal{C}_{i}^{\prime}$ are trained to maximize the target domain differences in each classifier pair. Third, with fixed $\mathscr{C}$, $\mathcal{F}$ is trained to minimize the target domain difference on each classifier pair. Fourth, for the target domain classification, the output would be the average of the $K$ classifiers driven from $K$ multiple sources.
\begin{figure*}[t!]
    \centering
    \begin{minipage}[c]{.24\linewidth}
    \includegraphics[width=\textwidth]{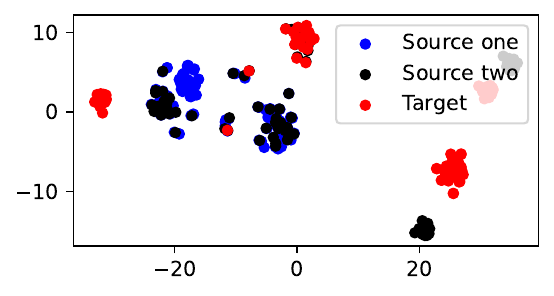}
    \end{minipage}
\hfill
    \begin{minipage}[c]{.24\linewidth}
    \includegraphics[width=\textwidth]{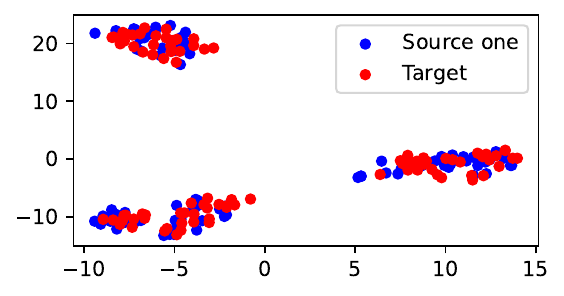}
    \end{minipage}
    \hfill
    \begin{minipage}[c]{.24\linewidth}
    \includegraphics[width=\textwidth]{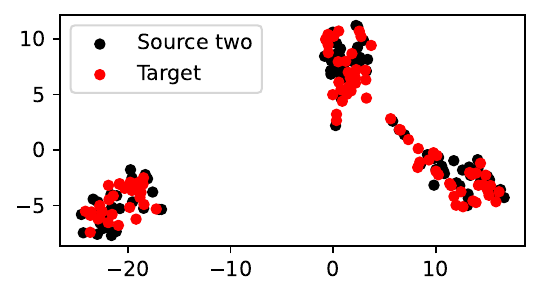}
    \end{minipage}
    \hfill
    \begin{minipage}[c]{.24\linewidth}
    \includegraphics[width=\textwidth]{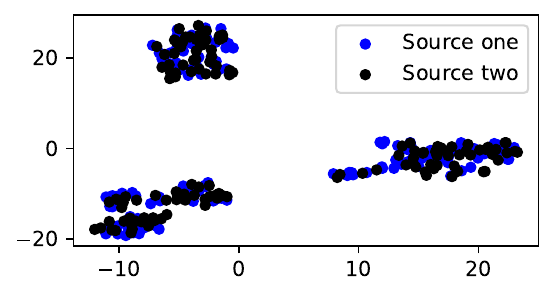}
    \end{minipage}
\begin{minipage}{0.24\textwidth}
 \centering
\text{(a)}
\end{minipage}
\begin{minipage}{0.24\textwidth}
 \centering
\text{(b)}
\end{minipage}
\begin{minipage}{0.24\textwidth}
 \centering
\text{(c)}
\end{minipage}
\begin{minipage}{0.24\textwidth}
 \centering
\text{(d)}
\end{minipage}
\vspace{-2 pt}
    \caption{t-SNE visualization of the proposed method (a) before adaptation, (b) alignment of source one and target, (c) alignment of source two and target, and (d) alignment of both sources.}
    \label{fig:tsne}
\end{figure*}
\section{Experimental Study}
\label{sec:results}
\subsection{Dataset}
\label{ssec: dataset}
In this study, we utilized the publicly available SEED dataset \cite{zheng2015investigating}. This dataset comprises $15$ movie clips that elicit happiness, sadness, and neutral emotional states. The dataset consists of 15 participants, comprising 8 females and 7 males. During the experiments, participants were instructed to fully immerse themselves in the movie clips to evoke the corresponding emotions. EEG signals were recorded using a total of $62$ channels and each trial adhered to a predefined sequence: a 5-second introductory hint, followed by $4$ minutes of the clip serving as the emotional stimulus, then $45$ seconds allocated for self-assessment, and finally a $15$-second break. EEG data were downsampled from $1000$ Hz to $200$ Hz, and a band-pass filter with a frequency range of $0.5$-$70$ Hz was applied. We calculate the differential entropy (DE) features at $1$-second intervals with no overlap in delta: $1-4$ Hz, theta: $4-8$ Hz, alpha: $8-13$ Hz, beta: $13-30$ Hz, and gamma: $30-50$ Hz frequency subbands.


\begin{table}[t]
\centering
\caption{Mean of the performance for the proposed method.  The “Source only” and “Target only” rows are the results on the target domain when using no domain adaptation and training only on the source or the target domain respectively.}
\begin{tabular}{lcc}

Method & Accuracy&F1-score \\
\hline
Source only&$0.86\pm 0.06$&$0.85\pm 0.07$ \\
\hline
Single Source&$0.88\pm 0.06$&$0.88\pm 0.06$ \\
MSDA-TF&$0.92\pm 0.04$&$0.92\pm 0.05$ \\
\hline
Target only&$0.95\pm 0.07$&$0.95\pm0.06$ \\
\hline
\end{tabular}
\label{table:proposed}
\end{table} 
\begin{figure}[t]
    \centering
    \begin{minipage}[c]{.23\textwidth}
    \includegraphics[width=\textwidth]{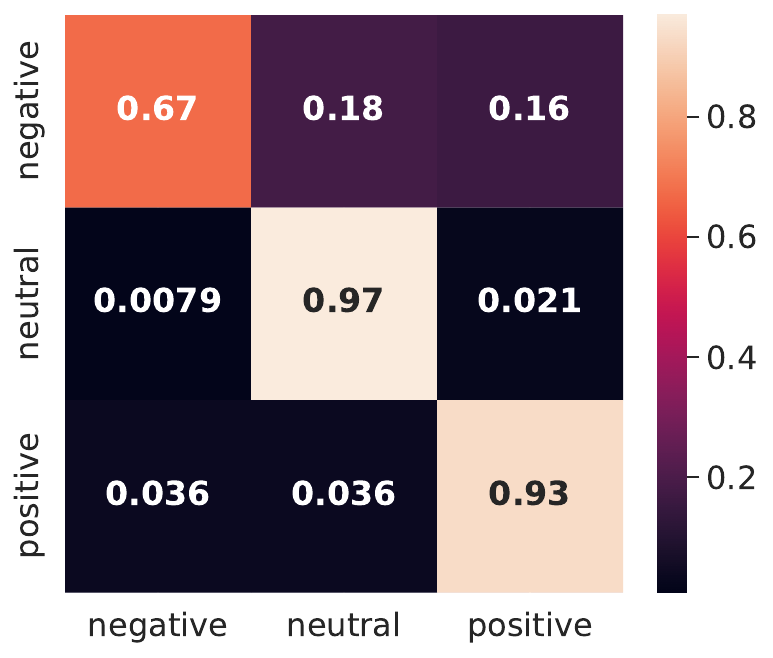}
    \end{minipage}
    \hfill
    \begin{minipage}[c]{.23\textwidth}
    \includegraphics[width=\textwidth]{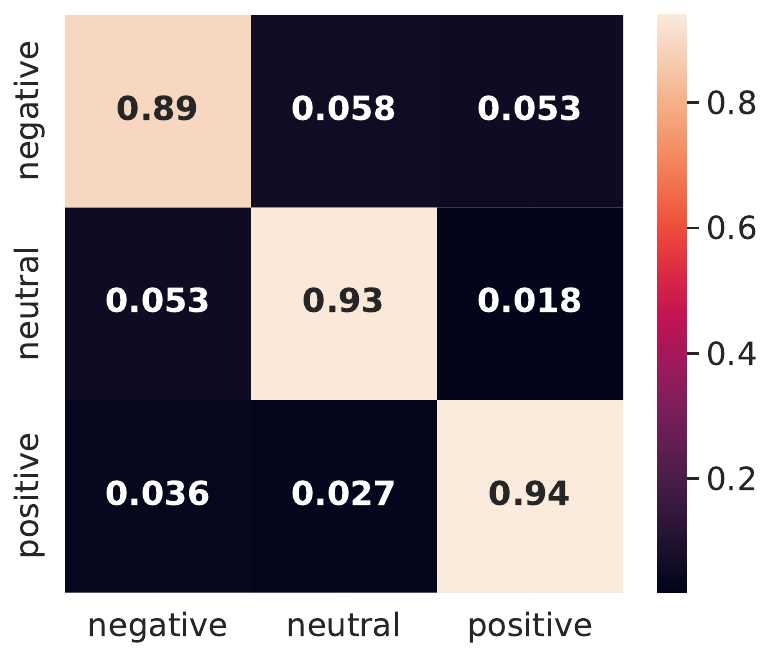}
    \end{minipage}
    \hfill
    \vspace{-1 pt}
    \caption{Confusion matrices for source-domain training with no adaptation (\textbf{left}), and proposed MSDA-TF approach (\textbf{right}).}
    \label{fig:confusion}
\end{figure}
\vspace{-1 pt}
\begin{table}[t]
\centering
\caption{Comparison with previous works.}
\begin{tabular}{lc}

Study & Accuracy \\
\hline
\textbf{Proposed MSDA-TF}&$\textbf{0.92}\pm \textbf{0.04}$ \\
Pan \textit{et al.} \cite{pan2023msfr}&$0.87\pm 0.05$ \\
She \textit{et al.}\cite{she2023multisource}&$0.86\pm 0.07$\\
Zhao \textit{et al.} \cite{zhao2021plug}&$0.86\pm 0.07$ \\
Du \textit{et al.}\cite{du2020efficient}&$0.90\pm 0.01$\\
\hline
\end{tabular}
\label{table:comparison}
\end{table}

\subsection{Results}
\label{ssec: results}
In this section, we present the performance of the proposed approach. The input data are normalized by subtracting the mean and dividing by the standard deviation. To conduct the evaluation, we adhere to the leave-one-subject-out cross-validation scheme, comprising 14 source subjects and 1 target subject in each validation round. Accuracy and F1-scores are calculated for each validation, and the average performance is reported. To group the source subjects, Pearson correlation is calculated among the source subjects and sorted in descending order. Based on the correlation scores, the subjects with the highest correlation values are divided into $4$ groups. Two groups of three subjects and two groups of four. The optimizer, learning rate, and number of epochs are set to Adam optimizer, $0.0001$, and $350$, respectively. 

Table~\ref{table:proposed} presents the performance results for the proposed MSDA-TF compared to three baseline methods, namely, Source only, Target only, and Single source, using the metrics of Accuracy and F1-score, along with their respective standard deviations. Training exclusively on the source domain (no domain adaptation) leads to an average accuracy of $0.86\pm 0.06$ and an F1-score of $0.85\pm 0.06$. Notably, when exclusively trained on the target domain, the model attains an average accuracy of $0.95\pm 0.07$, representing the highest performance achieved by the proposed feature generator. Moreover, we include the results of considering all source subjects as a single source domain. As presented, the proposed MSDA-TF results in an accuracy and F1-score of $0.92\pm0.04$ which highlights the positive effect of using domain adaptation with multiple source domains.

Furthermore, Figure~\ref{fig:confusion} displays the predictions for each emotion class as a confusion matrix. Comparing the results of using the source domain data with no adaptation with our proposed MSDA-TF approach, we observe that the proposed approach aids the model in detecting negative emotions significantly. Also, Table~\ref{table:comparison} presents a comparison of the proposed method with several methods from recent literature and demonstrates the superiority of our proposed approach.

To verify the MSDA process, we visualize the t-SNE \cite{vazquez2013virtual} of the learned representations corresponding to two source domains and a target domain before the classification step, as shown in Figure~\ref{fig:tsne}. While Figure~\ref{fig:tsne} (a) displays the scatter plot of the two source domains and the target domain, Figures~\ref{fig:tsne} (b-d) present the alignment of the target with each source and the sources with each other. This visualization suggests the proposed adaptation process works properly.   
\section{Conclusion}
\label{sec:conclusion}
In this work, we proposed a novel multi-source domain adaptation approach called MSDA-TF for subject-independent EEG-based emotion classification. Our method extracts feature representations from spectral, temporal, and spatial EEG characteristics and aligns the moments of the unlabeled target domain with each of the labeled source domains and the source domains with each other as well. The results demonstrate that MSDA-TF performs domain adaptation successfully and outperforms state-of-the-art algorithms.



\newpage
\bibliographystyle{IEEEbib}
\bibliography{strings,refs}

\end{document}